%% file: main.tex
\documentclass[12pt]{article}

\usepackage{arxiv}

\usepackage[utf8]{inputenc} 
\usepackage[T1]{fontenc}    
\usepackage[hidelinks]{hyperref}
\usepackage{url}            
\usepackage{booktabs}       
\usepackage{amsfonts}       
\usepackage{nicefrac}       
\usepackage{microtype}      
\usepackage{lipsum}
\usepackage[square,sort,numbers]{natbib}

\usepackage{inconsolata}
\usepackage{graphicx}
\usepackage{tabularx}

\usepackage{todonotes}

\makeatletter
\newcommand*\iftodonotes{\if@todonotes@disabled\expandafter\@secondoftwo\else\expandafter\@firstoftwo\fi}  
\makeatother

\usepackage{titlesec}

\titleformat{\section}
  {\normalfont\Large\bfseries}{}{0em}{}

\titleformat{\subsection}
  {\normalfont\large\itshape}{}{0em}{}

\titleformat{\paragraph}[runin]
  {\normalfont\normalsize\bfseries}{\theparagraph}{1em}{}
  [{.}]

\newcounter{boxed}
\newenvironment{boxed}[1][]{\begin{center}
\refstepcounter{boxed}
    \begin{tabular}{|p{\textwidth}|}
    \hline\\
    \textbf{Box~\theboxed. #1} \par\medskip
    }
    { 
    \\\\\hline
    \end{tabular} 
    \end{center}
    }

\title{Dissociating language and thought \\ in large language models}

\author{Kyle Mahowald*\\
The University of Texas at Austin\\
mahowald@utexas.edu \\
\And
Anna A. Ivanova*\\
Georgia Institute of Technology \\
a.ivanova@gatech.edu\\
\AND 
Idan A. Blank\\
University of California Los Angeles\\
iblank@psych.ucla.edu\\
\And
Nancy Kanwisher\\
Massachusetts Institute of Technology\\
ngk@mit.edu \\
\And
Joshua B. Tenenbaum\\
Massachusetts Institute of Technology\\
jbt@mit.edu\\
\And
Evelina Fedorenko\\
Massachusetts Institute of Technology\\
evelina9@mit.edu
}

\begin{document}

\maketitle

\begin{abstract}
    Large Language Models (LLMs) have come closest among all models to date to mastering human language, yet opinions about their linguistic and cognitive capabilities remain split. Here, we evaluate LLMs using a distinction between formal linguistic competence—knowledge of linguistic rules and patterns—and functional linguistic competence—understanding and using language in the world. 
    We ground this distinction in human neuroscience, which has shown that formal and functional competence rely on different neural mechanisms. Although LLMs are surprisingly good at formal competence, their performance on functional competence tasks remains spotty and often requires specialized fine-tuning and/or coupling with external modules. 
    We posit that models that use language in humanlike ways would need to master both of these competence types, which, in turn, could require the emergence of mechanisms specialized for formal linguistic competence, distinct from functional competence.
    \bigskip

\emph{* The two lead authors contributed equally to this work.}

\end{abstract}



\section{The language-thought conflation}
\label{introduction}

When we hear a sentence, we typically assume that it was produced by a rational, thinking agent (another person). The sentences that people generate in day-to-day conversations are based on their world knowledge (“Not all birds can fly.”), their reasoning abilities (“You’re 15, you can’t go to a bar.”), and their goals (“Would you give me a ride, please?”). Thus, we often use other people’s statements as a window into their minds.

In 1950, Alan Turing leveraged this tight relationship between language and thought to propose his famous test \cite{turing_computing_1950}. The Turing test uses language as an interface to cognition, allowing a human participant to probe the knowledge and reasoning capacities of two conversation partners to determine which of them is a human and which is a machine.
Although the utility of the Turing test has since been questioned, it has undoubtedly shaped the way society today thinks of machine intelligence \cite{chang2023survey}.

The popularity of the Turing test, combined with the language-thought coupling in everyday life, has led to several common fallacies related to the language-thought relationship. 
One fallacy is that an entity (be it a human or a machine) that is good at language must also be good at thinking. If an entity generates long coherent stretches of text, it must possess rich knowledge and reasoning capacities. Let’s call this the “good at language -> good at thought” fallacy. This fallacy has come to the forefront due to the recent rise of \textbf{Large Language Models} (LLMs; see Glossary), 
including OpenAI's GPT models, Anthropic's Claude, and more open alternatives \cite{bommasani2023foundation} like Meta's LLaMa models and EleutherAI's GPT-J.
LLMs today can produce text that is difficult to distinguish from human output, outperform humans at some text comprehension tasks \cite{wang2019superglue,srivastavaBIG}, and show superhuman performance on next-word prediction \cite{oh2023does}. As a result, claims have emerged—both in the popular press and in the academic literature—that LLMs are not only a major advance in language processing, but are also showing ``sparks of artificial general intelligence'' \cite{bubeck2023sparks}. 
However, when evaluating LLMs' capabilities, it is important to distinguish between their ability to {think} and their linguistic ability.
The ``good at language -> good at thought'' fallacy makes it easy to confuse the two, leading people to mistakenly attribute intelligence and intentionality to even the most basic dialog systems (e.g., the chatbot Eliza from the 1960s \cite{weizenbaum1966eliza}).

The contrapositive of this fallacy is that a model that is bad at thinking must also be a bad model of language. Let’s call this the ``bad at thought -> bad at language'' fallacy. LLMs are commonly criticized for their lack of consistent, generalizable world knowledge \cite{elazar-etal-2021-measuring}, lack of commonsense reasoning abilities \cite{marcus2020next}, and failure to understand what an utterance is really about \cite{bender-koller-2020-climbing}. Based on this evidence, some critics suggest that the models’ failure to produce linguistic output that fully captures the richness and sophistication of human \textit{thought} means that they are not good models of human \textit{language}.

Both the ``good at language -> good at thought'' and the ``bad at thought -> bad at language'' fallacies stem from the conflation of language and thought. This conflation is unsurprising: it is still novel, and thus uncanny, to encounter an entity that generates fluent sentences despite lacking a human identity. Thus, our heuristics for understanding what a language model is doing—heuristics that emerged from our language experience with other humans—are broken.

To mitigate the language-thought conflation fallacies, we propose to systematically distinguish between two kinds of linguistic competence: \textbf{formal linguistic competence}-the knowledge of rules and statistical regularities of language---and \textbf{functional linguistic competence}---the ability to use language in real-world situations. Our motivation for the formal/functional distinction comes from the human brain, where these skills are robustly dissociable. Both formal and functional linguistic competence are essential components of human language use: an effective communicator needs to both generate grammatical, meaningful utterances and strategically use those utterances to achieve diverse, context-dependent goals \cite{grice_logic_1975,clark1992arenas}.

Armed with this distinction, we evaluate the capabilities of contemporary LLMs and argue that LLMs exhibit a gap between formal and functional competence skills: for modern LLMs, formal competence in English is near human-level, whereas their functional competence remains patchy, with results depending on specific functional competence domains and on tasks within those domains.
Moreover, whereas formal linguistic competence in LLMs improves drastically with the amount of training data, functional linguistic competence improvements with scale are less consistent, such that LLM developers have now shifted away from simple scaling-up of the language prediction task to more specialized methods targeting behaviors of interest (e.g., \textbf{Reinforcement Learning from Human Feedback: RLHF} \cite{ouyang2022training}) or coupling an LLM with external specialized modules (leading to so-called ``augmented language models''; \cite{mialon2023augmented}). 

Therefore, we posit that the next-word prediction objective allows a model to master formal but not necessarily functional linguistic competence. 
What is required for mastery of functional competence is harder to pin down, in part because so much of human cognition (commonsense reasoning, scientific knowledge, everyday knowledge) can be conveyed \textit{through language} and thus learned \textit{from language}---even if these capacities are not themselves inherently linguistic.
As a result, language models acquire a variety of non-linguistic capacities.
But the ultimate ceiling for functional competence depends on important open questions about the information contained in the linguistic signal and the mechanisms used to leverage that information, as we discuss.

In the rest of the paper, we develop a framework for evaluating the competence of modern language models from a cognitive science perspective. In the first section, we elaborate on the constructs of formal and functional linguistic competence and motivate this distinction based on the evidence from human neuroscience. In the second section, we discuss the successes of LLMs in achieving formal linguistic competence, showing that models trained on word-in-context prediction capture numerous complex linguistic phenomena. In the third section, we consider several domains required for functional linguistic competence—formal reasoning, world knowledge, situation modeling, and social cognition—on which today’s LLMs often fail, or at least perform worse than humans. In the fourth section, we discuss the implications of our framework for building and evaluating future models of language and thought before summarizing our key conclusions in the final section.

\section{Formal vs. functional linguistic competence} \label{competence}

\subsection{What does linguistic competence entail?}

\paragraph{Formal linguistic competence}

We define formal linguistic competence as a set of capacities required to produce and comprehend a given language. 
Broadly, being formally competent means getting the \textit{form} of language right: knowing which strings could be valid words of a language (e.g., \textit{bnick} cannot be a word in English but \textit{blick} can \cite{halle1962phonology}), 
how to productively combine morphemes to form novel words (e.g., Barack Obama-less-ness but not Barack Obama-ness-less \cite{aronoff2022morphology}), learning enough about word meanings to know which words can go in which slots in a sentence \cite{cruse1986lexical}, and knowing how to combine words into valid sentences.

Because of its centrality in the history of linguistics, it is the last of these (forming words into sentences) that we focus on in our discussion of formal competence.
Most users of Standard Written English say, “The dogs in my bedroom are asleep” rather than “The dogs in my bedroom is asleep”, because the verb “to be” must match the number of the noun that is the subject of the sentence (“the dogs”), even though that verb is closer to an intervening, singular noun (“bedroom”). Linguistic competence also requires exquisite sensitivity to the regularitiesof idiosyncratic linguistic constructions. For instance, although English speakers know not to use the indefinite article ``a'' with plural nouns---making a phrase like ``a days'' ill-formed---they also know that it is allowed in a special construction where an adjective and a numeral intervene: “a beautiful five days in New York” \cite{dalrymple_amazing_2019,keenan_pleasant_2013}. 

Human language users likely learn rules, plus thousands of idiosyncratic constructions \cite{goldberg2019explain}, through some combination of sophisticated statistical learning \cite{bresnan2007syntactic,clark_distributional_2014,saffran1996statistical}
and innate conceptual and/or linguistic machinery \cite{chomsky1957syntactic,gleitman_human_1993,jackendoff2002foundations}. The result is the human ability to understand and produce grammatical and coherent linguistic utterances.

\paragraph{Functional linguistic competence}

In addition to being competent in the rules and statistical regularities of language, a competent language user uses language to accomplish goals in the world \cite{clark1996using,grice_logic_1975,bucholtz2004language}: to talk about things that can be seen or felt or heard, to reason about diverse topics; to make requests; to cajole, prevaricate, and flatter. People use language in tandem with other perceptual and cognitive systems, such as our senses and our memory, and deploy words as part of a broader communication framework supported by our sophisticated social skills. A formal language system in isolation is useless unless it can interface with the rest of perception, cognition, and action.

\begin{figure}
    \centering
    \includegraphics[width=\textwidth]{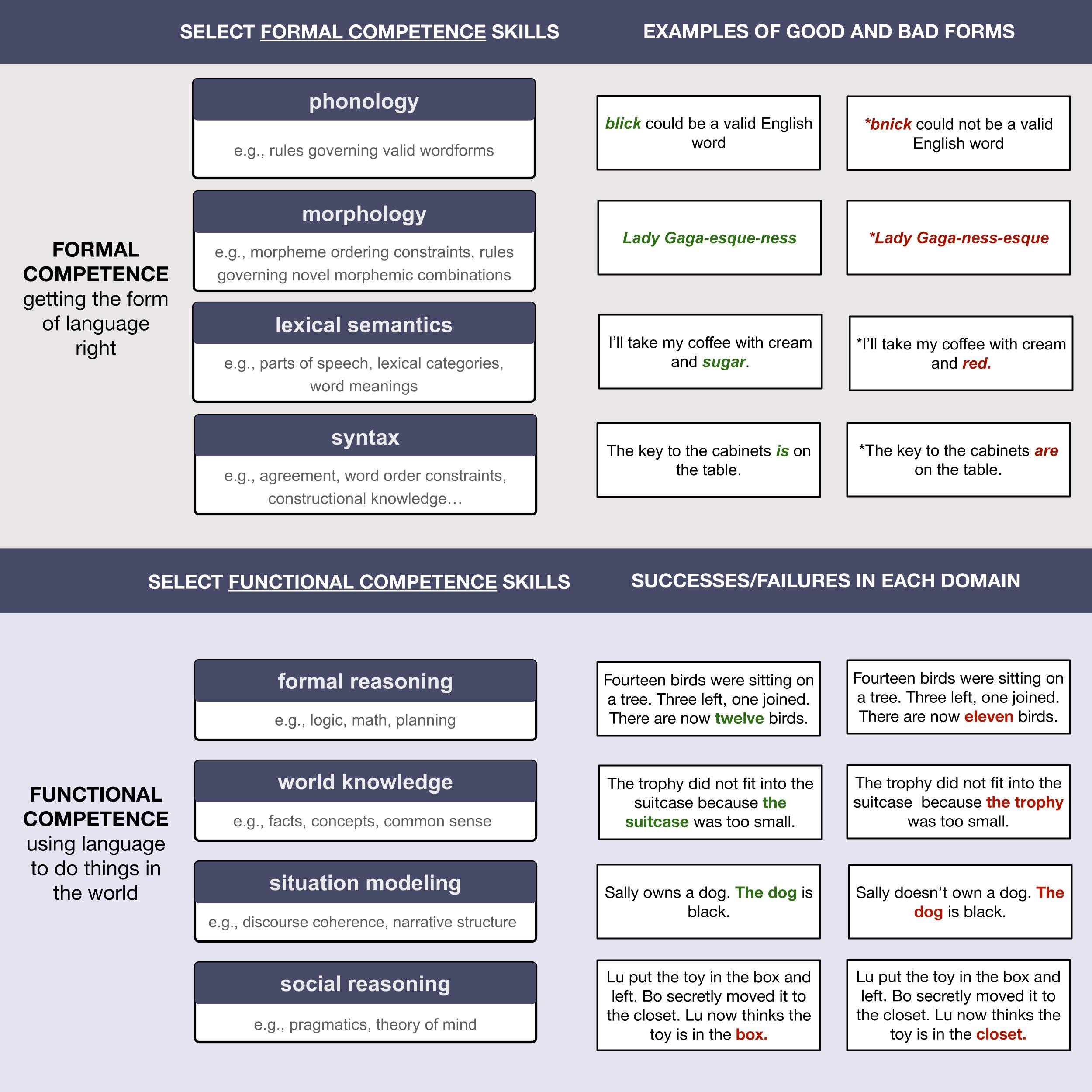}
    \caption{\textbf{Separating formal and functional competence.} Successful use of language relies on multiple cognitive skills, some of which (required for formal competence) are language-specific and some (required for functional competence) are not.
    Determining whether a particular failure stems from a gap in formal competence or functional competence is key to evaluating and improving language models.}
    \label{fig:fig1}
\end{figure}

The capacities required to use language to do things in the world are distinct from formal competence and depend crucially on non-linguistic cognition (Figure \ref{fig:fig1}). Thus, we define functional linguistic competence as non-language-specific cognitive functions that are required when using language in tandem with non-language-specific capacities in real-world circumstances.

\subsection{Motivation for the distinction between formal vs. functional linguistic competence}

Our motivation for the distinction between formal and functional linguistic competence comes from what we know about the architecture of the human mind. In humans, language is robustly dissociated from the rest of high-level cognition, as well as from perception and action. Below we briefly summarize a body of evidence from cognitive science and neuroscience that supports this dissociation.

\paragraph{The language network supports language processing in the human brain} \label{language_network_introduction}

Human language processing draws on a set of interconnected brain areas in the frontal and temporal lobes (typically in the left hemisphere). This \textbf{language network} supports both comprehension (spoken, written, and signed) \cite{deniz_representation_2019, fedorenko_new_2010, macsweeney_neural_2002, scott_new_2017} and production \cite{menenti_shared_2011, hu2023precision}; 
is sensitive to linguistic regularities at multiple levels: from phonological/sub-lexical \cite{regev} to phrase/sentence level \cite{fedorenko_functional_2011, fedorenko_lack_2020-1}; and supports linguistic operations related both to the processing of word meanings and to combinatorial semantic and syntactic processing \cite{fedorenko_lack_2020-1, hu2023precision}. Damage to the language network leads to linguistic deficits \cite{bates_voxel-based_2003, wilson_language_2019}. This tight link between the language network and language function indicates that these brain regions are responsible for language processing in humans.

\paragraph{The language network does not support non-linguistic cognition} \label{language_network_selectivity}

The language network is remarkably selective for language. Evidence of a strong dissociation between language processing and non-linguistic abilities comes from two main sources: a) functional brain imaging studies of neurotypical adults, and b) behavioral investigations of individuals with aphasia---a language impairment typically caused by a stroke or degeneration.

Brain imaging techniques like functional MRI (fMRI) are used to observe real-time activity in the language network in healthy individuals. Given its high spatial resolution, fMRI is well-suited to study whether any two cognitive abilities draw on the same brain structures. For example, to ask whether language and mathematical reasoning recruit the same brain areas, we can have participants perform a language task and a math task while in an MRI scanner and then test whether brain regions that are active during language processing are also active when participants solve a math problem. This approach reveals that the language network is extremely selective for language processing: it responds reliably when people listen to, read, or generate sentences, but not when they perform arithmetic tasks, engage in logical reasoning, understand computer programs, listen to music, categorize objects or events, reason about people’s mental states, or process non-verbal communicative information like facial expressions or gestures \cite{amalric_origins_2016, benn2023language, chen, deen_functional_2015, fedorenko_functional_2011, jouravlev_speech-accompanying_2019, liu2020computer, monti_thought_2012, paunov_differential_2022}. 

Studies of individuals with aphasia provide a unique opportunity for testing which cognitive capacities rely on linguistic representations. Of particular interest are cases of `global aphasia', which affects both production and comprehension. Individuals with global aphasia exhibit severe linguistic deficits that spare nothing but a small set of words. If some aspects of non-linguistic cognition draw on the same resources as language, then individuals with severe linguistic deficits should invariably exhibit impaired performance on the relevant non-linguistic tasks. However, despite the nearly complete loss of linguistic abilities, individuals with severe aphasia can have intact non-linguistic cognitive abilities: they can play chess, solve arithmetic problems, leverage their world knowledge to perform diverse tasks, reason about cause and effect, and navigate complex social situations \cite{fedorenko_language_2016}.

In summary, evidence from brain imaging studies and from individuals with aphasia is remarkably consistent: the mechanisms that process language in the human brain do not support non-linguistic cognitive tasks. This sharp dissociation suggests that, in examining language models’ functionality, we should separate their linguistic abilities from their abstract knowledge and reasoning abilities, which can be probed—--and perhaps even learned--—through a linguistic interface, but which require more than formal linguistic competence.

\section{LLMs have largely mastered formal linguistic competence in English} \label{formal_competence}

In a 2019 interview, Chomsky remarked \cite{fridman2019chomsky}: “We have to ask here a certain question: is [deep learning] engineering or is it science? […] On engineering grounds, it’s kind of worth having, like a bulldozer. Does it tell you anything about human language? Zero.” The view that deep learning models are not of scientific interest remains common in linguistics, and, despite many arguments for integrating such models into research on human language processing and acquisition \cite{linzen2019can,blank2023large,jain2023computational} and a chorus of arguments that they should be taken seriously as linguistic and cognitive models \cite{frank2023openly,cao2021explanatory,baroni2022proper}, their integration into language research still encounters resistance.

In this section, we evaluate the performance of LLMs \textit{qua} language models by asking whether these models have made progress towards achieving formal linguistic competence--the kind of competence supported by the language-selective network in the human brain. 
We argue that LLMs have turned out to be surprisingly successful at mastering formal competence---qualitatively different in their formal linguistic capacities from models from before roughly 2018 in a way that was predicted by few practitioners in the field, and which was unexpected given longstanding claims that grammatically competent systems would require strong language-specific priors.
With surprise comes information: models' successes are informative for linguistic theorizing.

\subsection{Statistical language models: some fundamentals}

LLMs arose from several earlier approaches in computational linguistics, including statistical language modeling, word embeddings, and connectionism (an earlier term for the approach that morphed into today’s deep learning). Similar to earlier statistical language models, LLMs are usually first trained on a word prediction task (the same task used for training n-gram models going back to Shannon’s work in the mid-20th century; see \cite{jurafsky2009speech} 
for historical overview). Similar to approaches in distributional semantics and word embeddings (for overviews, see \cite{baroni2010distributional,erk2012vector}), LLMs represent linguistic information as vectors in a high-dimensional space. 
Similar to earlier connectionist approaches \cite{rumelhart_parallel_1986,elman1993learning}, LLMs are neural networks --- a class of machine learning systems that was originally inspired by the human brain and learns its parameters from the input data. All of these approaches stand in contrast to models that use explicit, structured hierarchical representations of syntactic rules (see \cite{norvig2012colorless} for a discussion of these two divergent paradigms).

N-grams and word embedding models achieved some success in various domains in natural language processing (e.g., spelling correction, spam classification, sentiment analysis).  However, they never approached human-level performance on general language tasks like text generation, leading to claims that purely statistical approaches would never be able to capture the richness of natural language, particularly in complex syntactic, morphological, and semantic domains \cite[e.g.,][]{pinker_language_1988}. For instance, it has been claimed that statistical approaches, which use linear strings of words as input, are in principle unable to learn rare and complex syntactic features that require representing phrases and sentences hierarchically \cite{everaert2015structures}. This pessimism is now challenged by LLMs.

LLMs are typically first trained on a training set constructed from a massive amount of text from the web. During \textbf{pretraining}, LLMs have a simple objective: predict a held-out \textbf{token} (the basic unit in LLMs---often but not always corresponding to words or morphemes \cite{sennrich-etal-2016-neural}) based on a fixed number of previous tokens.
The predicted token is then compared with the ground truth (which token actually occurred in that sentence), and the error signal is propagated back through the model to update its many parameters.
The token prediction objective is often used as a pretraining step, and the model is then \textbf{fine-tuned} for a more specific task.

Although it is tempting \cite{bowman-2022-dangers} to move the goalposts and focus on what these models are still unable to do, we argue that the remarkable advances in LLMs’ ability to capture various linguistic phenomena should not be overlooked. 
Significant formal linguistic abilities arise in models on the scale of GPT-2 or BERT and seem to plateau at a high level in contemporary LLMs (Box \ref{box:trajectory}).

\begin{boxed}[The path toward formal linguistic competence] \label{box:trajectory}
When did LLMs achieve formal linguistic competence? Table I shows text generations from an n-gram model, an RNN (a recurrent neural network, which were state-of-the-art language models in the mid-2010s, before transformers), GPT-2, GPT-3, and GPT-4. 

\vspace{.14in}

The n-gram model and RNN clearly fail to show formal linguistic competence: the sentences are not grammatical or sensible.
In contrast, GPT-2 (a model produced just 8 years after the RNN in Row 2) generates grammatical output: the verbs agree with their subjects, parts of speech are used appropriately, relatively complex syntactic constructions like the relative clause ``that has nothing to do with who we are'' is appropriately tied to a noun phrase. 
However, compared to the GPT-3 and GPT-4 output, it makes less sense: what does ``everything about being with people'' mean? 
Finally, the outputs of GPT-3 and GPT-4 are both grammatical and semantically coherent, such that they can plausibly be taken for answers generated by a human. 

\vspace{.14in}

\textbf{Table I}. \emph{Examples of texts generated by increasingly more advanced language models in response to a prompt ``The meaning of life is...'' (following \cite{sutskever2011generating} and using their example verbatim in Row 2). Starting with GPT-2, outputs are grammatical and largely use the rules and patterns of English (formal linguistic competence).}

\vspace{.07in}

  \begin{tabularx}{\textwidth}{|l|X|}
      \hline\smallskip
    \textbf{Model} & \textbf{Text} \\ \hline\smallskip
    3-gram model trained on Moby Dick & \textbf{\emph{The meaning of life is}} gulped and gone, bleeds with keenest anguish at the same muteness of humanity over all creations. Just so with Ahab, 'says I;' look here you sabbee me, has every reason to be served.
 \\ \hline\smallskip
    RNN from \cite{sutskever2011generating} & \textbf{\emph{The meaning of life is}} the tradition of the ancient human reproduction: it is less favorable to the good boy for when to remove her bigger. In the show’s agreement unanimously resurfaced.
 \\ \hline\smallskip
     GPT-2 \cite{radford_language_2019} & \textbf{\emph{The meaning of life is}} something simple that has nothing to do with who we are. Everything about being with people is tied to those characteristics, and people are just machines. \\ \hline
     GPT-3 davinci-002 \cite{brown2020language} & \textbf{\emph{The meaning of life is}} a mystery to us all, and a question that will never be answered. Life is what we make of it, and it's up to each of us to find our own meaning in life. \\ \hline
     GPT-4 (via ChatGPT interface) & \textbf{\emph{The meaning of life is}} subjective and varies greatly depending on personal beliefs, values, and experiences. Some people find meaning in relationships, personal achievements, or spiritual beliefs, while others may see it as a journey of self-discovery, learning, or contributing to the greater good. \\\hline
  \end{tabularx}
           \label{tab:models}
\end{boxed}

\begin{boxed}[What about semantics?] \label{box:semantics}
    Does semantics fall under formal or functional linguistic competence? The answer depends on what kind of \emph{semantics} we are talking about. 

    \vspace{.14in}

    One meaning of semantics, often associated with compositional and lexical semantics, concerns the way that meaning is derived from words and their combinations. We consider this aspect part of formal linguistic competence. Indeed, the language network in the brain responds both to lexical semantics, i.e., retrieving the meaning of individual words, and to compositional semantics, i.e., constructing the meaning of multi-word utterances \cite{fedorenko_lack_2020-1}. Insofar as LLMs are highly sensitive to lexical and combinatorial semantics, they parallel the language network.

    \vspace{.14in}

    The second meaning of semantics is something closer to``general conceptual knowledge'' (used in contexts such as ``non-verbal semantics'', e.g., extracting the meaning of a picture). This definition is closely related to the notion of world knowledge, which we discuss in our section on world knowledge. Given the fact that conceptual knowledge and reasoning does not \emph{have} to operate over linguistic inputs but is nonetheless essential for fluent language use, we classify it under functional linguistic competence. 
\end{boxed}

\subsection{Large language models learn core aspects of human language processing} \label{hierarchy_and_abstraction}

For LLMs to be useful as models of language processing in humans, we must be convinced that the models encode the abstract phonological, morphological, syntactic, and semantic rules that characterize human language (see Box \ref{box:semantics} for a distinction between ``linguistic'' and conceptual semantics). Although interesting differences exist between linguistic processing in LLMs and humans \cite{lenci2023understanding,van2021single}, there are also important similarities.
Here, we review evidence that LLMs succeed as models of formal linguistic competence.
We focus primarily on syntax, showing evidence of mastery on grammatical benchmarks as well as evidence for emergent syntactic structure in LLMs.
But similar successes, both in performance and emergent structure, have been shown in other linguistic domains (e.g., emergent phonological structure \cite{beguvs2021ciwgan}, productive generation of morphologically complex neologisms \cite{mccoy-etal-2023-much}, rich lexical semantic information \cite{chronis-erk-2020-bishop}, etc.).

\paragraph{LLMs perform well on benchmarks of diverse linguistic phenomena}

By being trained for word prediction, transformer models learn a lot about the structure of language, including linguistic features that, even recently, were thought to be beyond the scope of statistical models. These models have succeeded not just on tests of general language understanding developed by the NLP community (e.g., GLUE tasks \cite{wang2018glue}), but, critically for our purposes, on tests of linguistic competence in English and other languages with massive corpora available (see Box 3 for discussion of lower-resourced languages). 

The benchmark BLiMP \cite{warstadt2020blimp}, for instance, contains minimal pairs of grammatical vs. ungrammatical sentences across a diverse range of complex linguistic phenomena like filler-gap dependencies (``Bert knew what many writers find'' vs. ``*Bert knew that many writers find'') and negative polarity items (``The truck has clearly tipped over'' vs. ``*The truck has ever tipped over'').
Strikingly, a model \cite{samuel-2023-mean} submitted to the BabyLM challenge \cite{warstadt-etal-2023-findings} achieved 86\% on BLiMP (cf. human baseline of 89\%) despite being trained on an amount of data comparable to what a human child might be exposed to (see Box 3).
Models achieve similarly impressive results on other linguistic benchmarks like SyntaxGym \cite{gauthier2020syntaxgym}, and there are now dozens of investigations of specific complex linguistic phenomena (some of which we discuss below).

\paragraph{LLMs learn hierarchical structure}

In human languages, words are combined to make compositional meanings. In a multi-word sentence, the individual words' meanings do not simply get added linearly one by one. Instead, they can be combined hierarchically into tree-like structures.

The \textbf{hierarchical structure} in language manifests in many ways. One prominent example is non-local feature agreement. In English and many other languages, verbs agree with their subjects. For instance, a plural subject uses the verb “are”, whereas a singular subject uses “is”.
A bigram model, which simply stores frequencies of two-word strings, could learn that “The keys are on the table” is more probable than “The keys is on the table” by knowing that “keys are” is more common than “keys is”. But such a model would not be able to learn that the subject and verb agree even if arbitrarily far apart: for instance, “The keys to the old, wooden kitchen cabinet are on the table” has six intervening words between the subject and verb, and yet “are” still agrees with “keys” and not with “cabinet”. However, a model that learns the underlying hierarchical structure of English should be able to keep track of this long-distance subject-verb dependency \cite{linzen-etal-2016-assessing}. 

Today’s LLMs perform long-distance number agreement well above chance, preferring the grammatical over a non-grammatical sentence continuation even in the presence of intervening distractor words \cite{gulordava-etal-2018-colorless,l2021}, although some earlier models can be distracted by frequency effects (such as differences in the frequency between the singular and plural forms \cite{yu-etal-2020-word}).
In a similar vein, LLMs can handle other constructions that require complex hierarchy, like filler-gap dependencies \cite{wilcoxLI}. Finally, studies that examine the internal geometry of the models' sentence representations \cite{hewitt-manning-2019-structural}, studies that causally intervene on models' internal representations \cite{ravfogel-etal-2021-counterfactual}, and studies that turn on and off specific model ``neurons'' \cite{mueller-etal-2022-causal,lakretz-etal-2019-emergence} have provided mechanistic insights into how an LLM might represent hierarchical structure and establish non-local structural dependencies.

\paragraph{LLMs learn linguistic abstractions} \label{abstractions}

Following \cite{ambridge2020against}, we define an \textbf{abstraction} as a generalized linguistic representation—--such as a part-of-speech category (e.g., noun or verb) or grammatical role (e.g., subject or object)--—that goes beyond simple storage of input and allows for generalization. The very notion of subject-verb agreement, outlined in the previous section, relies on the abstract categories of subject and verb. As described in \cite{gulordava-etal-2018-colorless}, in a sentence like ``dogs in the neighborhood often… (bark/barks)'', a model might learn a shallow version of the agreement rule, namely, that the collocation of “dogs” and “bark” in the same sentence is more common than “dogs” and barks”. However, a model that has an abstract representation of categories like grammatical subject, grammatical number, and verb should be able to handle long-distance number agreement even for novel combinations of words. 

One way to test a model’s knowledge of abstract rules is by using semantically nonsensical sentences, like “The colorless green ideas I ate with the chair… (sleep/sleeps)”. 
Models have been shown to perform the agreement task well in several languages, even on these semantically anomalous sentences \cite{gulordava-etal-2018-colorless}.

An even more stringent test for linguistic abstraction asks whether LLMs can apply morphosyntactic rules to novel words. A study of BERT's abstraction capabilities  \cite{kim-smolensky-2021-testing} showed that BERT has some ability to generalize grammatical categories. They give the model novel words, used in phrases, as input (e.g., “the blick” where blick is likely a noun and “they dax” where dax is likely a verb) and test whether, based on the input, the model can generalize the part-of-speech category (e.g., assign a higher score to “I went to a blick” than to “I went to a dax”). They conclude that BERT succeeds partially at this task: it does learn to generalize, but only after repeated examples \cite[but see][for ways in which the word itself affects compositional ability]{kim2022uncontrolled,misra-etal-2023-comps}. Models also seem to (often) be able to use novel words appropriately \cite{brown2020language,mccoy-etal-2023-much}.

A large body of work has tested linguistic abstraction in LLMs using a method called probing \cite{ettinger-etal-2016-probing,belinkov-2022-probing}. In this literature, a classifier is often trained to take as input internal model representations and then predict as output an abstract category, such as part-of-speech or dependency role.
The logic of the probe is to test whether these abstract categories can be successfully recovered from the internal model states.
 Using this approach, it has been claimed that LLMs ``rediscover the classical NLP pipeline'' \cite{tenney-etal-2019-bert}, learning at various layers features like part-of-speech categories, parses, named entities, and semantic roles (although see  \cite{niu-etal-2022-bert}).

Importantly, a human-like language model is not expected to rely solely on abstract rules. 
Humans use diverse cues in their language learning and processing that sometimes override or conflict with strict hierarchical syntactic processing \cite[e.g.,][]{macdonald_lexical_1994,
bates}.
Humans also rely, to varying extents, on memorizing previously seen input, as opposed to purely applying abstract rules \cite{ambridge2020against,goldberg2019explain}.
Thus, when evaluating formal competence in LLMs, it is essential to directly compare their performance with that of humans \cite{dasgupta2022language}. For instance, a re-examination \cite{lampinen2022can-recursion} of an earlier study \cite{lakretz_causal_2021} showed that apparent syntactic agreement deficits in GPT-2 occurred on instances that were also challenging for humans. Overall, LLMs clearly learn some linguistic abstraction, even if the degree of that abstraction remains a matter of debate (as it does for humans).

\paragraph{LLMs learn constructions}

Recent evidence suggests that LLMs learn syntactic {constructions} \cite{weissweiler-etal-2023-construction,
tseng-etal-2022-cxlm,tayyar-madabushi-etal-2020-cxgbert}. These constructions can be idiosyncratic, lexically sensitive, and relatively rare, such as ``a beautiful five days in Austin'' \cite{mahowald-2023-discerning}. 
LLMs also show some amount of sensitivity to the Preposing in Prepositional Phrase construction (``Surprising though it may be...''), even when the gap crosses a finite clause boundary (``Surprising though \textit{I know} it may be'') \cite{potts2023}. 
They achieve this sensitivity even though such examples crossing the finite clause boundary are vanishingly rare: only 58 examples out of  {7 billion} sentences in a corpus.
The fact that models can learn that some vanishingly rare constructions are grammatical,  whereas other equally rare constructions are not, suggests that LLMs meaningfully learn something about syntax.

Models are also sensitive to the form of the comparative correlative ``the better the syntax, the better the semantics'' \cite{weissweiler-etal-2022-better}.
However, this sensitivity does not mean that they are sensitive to the semantic implications of the construction. 
Indeed, it appears that inferences based on these sentences can be a challenge (e.g., knowing that if I say ``the better the syntax, the better the semantics'' and then tell you that the syntax is better, this means the semantics is better).
This asymmetry nicely illustrates the formal/functional distinction: the model clearly knows how to use the construction and get the form right without necessarily being able to access the intended meaning.
We discuss these issues in more detail in later sections.

\subsection{LLMs are predictive of activity in the human language network} \label{llms_vs_languagenetwork}

As discussed above, language processing in humans relies on a dedicated brain network. This network exhibits all the hallmarks of formal linguistic competence: it is sensitive to abstract hierarchical rules in isolated phrases and sentences \cite{fedorenko_new_2010, fedorenko_neural_2016, pallier_cortical_2011, law_lists_2021}, in naturalistic narratives \cite{shain_fmri_2020, brennan_localizing_2020, heilbron_hierarchy_2022, reddy_can_2021}, and in syntactically well-formed but semantically empty ("jabberwocky") stimuli \cite{fedorenko_new_2010, fedorenko_language_2016, pallier_cortical_2011}. The language network is also sensitive to specific word co-occurrences \cite[e.g., as evidenced by sensitivity to n-gram surprisal; ][]{shain_fmri_2020}, indicating that it learns not only the rules, but also the patterns of language. The language network's selectivity for linguistic vs. non-linguistic inputs, along with its sensitivity to linguistic rules and patterns, allows us to operationalize formal linguistic competence as a set of computations that in humans take place within the language network.

If LLMs and the human language network perform similar computations to achieve formal linguistic competence, we expect to observe similarities in their internal organization (see \cite{cao2021explanatory} for similar arguments in the domain of vision). And indeed, LLMs and the human language network exhibit non-trivial similarities. 

First, the internal architecture of LLMs resembles that of the language network. Both operate at the level of abstract linguistic units (words/tokens) rather than modality-specific representations, such as pixels or acoustic waveforms, and combine these unit-level representations into composite representations of phrases and sentences. Neither system shows clear spatial segregation for syntactic and semantic processing (LLMs:  \cite{tenney-etal-2019-bert, huang-etal-2021-disentangling}; brain: \cite{fedorenko_lack_2020-1, reddy_can_2021}), indicating that these processes are tightly functionally coupled in both.

Second, one can establish a direct mapping between internal LLM representations and neural activity patterns within the language network. This mapping can be successfully used to predict brain responses to novel sentences and words in previously unseen contexts \cite{caucheteux_brains_2022, goldstein_shared_2022, schrimpf_neural_2021}. This similarity between sentence activation patterns in LLMs and the brain is suggestive of similar representational mechanisms that support computations in these systems.

We do not claim that the correspondence between LLMs and the language network is one-to-one.
For instance, LLMs learn patterns outside traditional human linguistic competency, such as predicting newline characters \cite{michaud2023quantization}. Nevertheless, the fact that internal representations learned by contemporary LLMs contain sufficient information to predict the language network's responses to diverse linguistic strings indicates at least some correspondence between LLMs' representations and those in the language network.

\subsection{Using LLMs as models of formal linguistic competence in humans}

LLMs today generate highly coherent, grammatical texts that can be indistinguishable from human output. In doing so, they exhibit knowledge of hierarchical structure and linguistic abstractions, while resembling human brain responses during language processing. These models are not perfect learners of abstract linguistic rules, but neither are humans. We therefore conclude that LLMs possess substantial formal linguistic competence, at least in English.

LLMs have already overturned claims about the fundamental impossibility of acquiring certain linguistic knowledge---including hierarchical structure and abstract categories---from the statistics of linguistic input alone \cite{piantadosi2023modern}. If language modeling continues to improve (including learning from more realistic kinds and amounts of data; Box \ref{box:limitations}), this would allow testing more general versions of this ``poverty of the stimulus'' argument \cite{chomsky_linguistics_1991}, including specific tests of what inductive biases might be necessary to successfully learn the rules and statistical regularities of human language. As such, LLMs have substantial value in the scientific study of language learning and processing.

\begin{figure*}

\begin{boxed}[Limitations of LLMs as human-like language learners and processors] \label{box:limitations}

A preponderance of evidence suggests that LLMs acquire formal linguistic competence. Here, we address three common criticisms of LLMs as models of human language processing. 

\subsection*{Excessive reliance on statistical regularities} \label{limitation_data}

LLMs succeed, in part, by learning statistical regularities, and they can be ``right for the wrong reason'' \cite{mccoy-etal-2019-right}. Many of LLMs' successes can be explained by patterns present in training data, such that a slight deviation from these patterns can drastically decrease performance \cite{mccoy2023embers}. In particular, adding noise or distracting information can degrade model performance \cite{kassner-schutze-2020-negated,misra-etal-2023-comps}
in ways that do not always affect humans. 
These results raise the related question of whether LLMs are just storing and regurgitating their inputs. But a study of GPT-2's novel behavior \cite{mccoy-etal-2023-much} showed that, although n-grams up to length 4 often appeared in the training set, GPT-2 generated mostly novel 5-grams and above.  Thus, LLMs seem to be capable of some meaningful generalizable morphosyntactic knowledge beyond mere memorizing.

\subsection*{Unrealistic amounts of training data}

Today's best LLMs are trained on vastly more data than a child is exposed to \cite{warstadt}, and some evidence suggests that a model’s training dataset would need to be unrealistically large to handle some constructions in a human-like way \cite{van-schijndel-etal-2019-quantity}, without stronger priors \cite{mccoy-etal-2020-syntax, yedetore-etal-2023-poor}.
This difference in the amount of input that models vs. human language learners require is sometimes taken to imply that the models will necessarily be un-humanlike. 
However, there is reason for optimism.
The BabyLM competition \cite{warstadt-etal-2023-findings} solicits submissions of language models trained on either a fixed corpus of 10M or 100M words---which are posited to be in the range of the number of words heard by a 10-year-old child.
The best-performing model in 2023 \cite{georges-gabriel-charpentier-samuel-2023-layers} achieved impressive performance on the BLiMP benchmark for syntactic minimal pairs, suggesting the possibility that models can learn grammar from smaller amounts of data (see also \cite{wilcoxLI}, among others). 
It has also been shown that smaller models still provide good matches to human neural responses \cite{hosseini2022artificial}.

\vspace{.1in}

Although performance of smaller language models is not perfect, improvements in language models—including the use of more cognitively-inspired architectures and learning algorithms---could lead to strong performance with far less training data.
As such, important questions are: what inductive biases are introduced by the model architectures and can those biases be made more human \cite{mccoy-etal-2020-syntax}?

\subsection*{Insufficient tests on languages other than English}

Because LLMs are data-hungry, they work best on languages for which vast corpora are available. For most human languages, this is not the case. Moreover, the architectures themselves may be biased towards English and other European languages \cite{blasi-etal-2022-systematic}, and not all languages are equally easy to model given existing infrastructure \cite{mielke-etal-2019-kind}.
That said, evidence is growing of strong performance in a variety of languages \cite{martin-etal-2020-camembert}, 
and successful transfer of models to low-resource languages \cite{wang-etal-2020-extending}.
We should still, however, proceed with caution in assuming that the success of LLMs will extend to all languages.
This is of particular concern for languages that are typologically distinct from English or have a different modality (e.g., signed languages).
\end{boxed}

\end{figure*}

\section{Non-augmented LLMs fall short on functional linguistic competence} \label{functional_competence}

Real-life language use is impossible without non-linguistic cognitive skills. Understanding a sentence, reasoning about its implications, and deciding how to respond all rely on cognitive capacities that go beyond formal competence. In this section, we ask: how good are contemporary LLMs at functional linguistic competence?

We focus on four key capacities that are not language-specific but are nevertheless crucial for language use in real-life settings: {i) formal reasoning}---a host of abilities including logical and mathematical reasoning, computational thinking, and novel problem solving; {ii) world knowledge}---factual and commonsense knowledge about agents, objects, properties, actions, events, and ideas; {iii) situation modeling}---the dynamic tracking of objects, agents, and events as a narrative/conversation unfolds over time; and {iv) social reasoning}---understanding the social context of linguistic exchanges. An average conversation requires the use of all these capacities, yet none of them are specific to language use. 

For each domain, we first describe its neural mechanisms in humans and then discuss how well contemporary LLMs have mastered the domain. We conclude that, unlike formal competence, functional competence of LLMs is uneven, often requiring specialized fine-tuning and/or lacking human-like robustness and generality. 
In Box \ref{box:methodology}, we highlight the importance of properly evaluating LLMs; evaluation issues can occur in studies of either formal or functional competence, but we believe they have led to a particularly large amount of overclaiming about models' functional competence.

\begin{figure*}[!t]
\begin{boxed}[Important considerations for evaluating functional competence] \label{box:methodology}

In discussing different domains of functional competence, it is important to highlight two key considerations.

\subsection*{A. Fine-tuning on the task and the challenge of closed models}

When discussing whether an LLM excels in a particular domain, it is essential to note whether the model has been fine-tuned on the task of interest. Formal competence skills can be observed in many LLMs trained on word-in-context prediction, without the need to specifically fine-tune them on syntactic trees or other grammatical abstractions. Functional competence skills, however, are often boosted through additional fine-tuning on some relevant corpus, the task of interest, or using more general fine-tuning techniques like reinforcement learning based on human feedback (RLHF). Claims such as ``LLMs succeed at Theory of Mind'' often apply to fine-tuned models, not the generic word-in-context prediction machines. 

\vspace{.14in}

An extreme case of task-specialized fine-tuning is when the task materials are, in fact, included in the model's training set. LLMs can memorize large amounts of information (and this trend becomes more pronounced in larger models; \citealp{tirumala2022memorization}), such that prompting them with the exact sentences (or similar sentences) they have encountered during training will give highly inflated performance metrics.

\vspace{.14in}

Identifying whether performance boost on a given task comes from generic changes to the model (such as increasing the model size or the amount of training data) vs. task-specific changes (such as fine-tuning on a similar task) is essential for further model understanding and improvement.

\vspace{.14in}
The closed nature of many LLMs (like the latest GPT models) make evaluating them in these ways difficult or impossible. 
Thus, we believe that useful scientific knowledge will increasingly come from studying models whose training data, architecture, and training procedure are transparent and able to be studied \cite{frank2023openly,bommasani2023foundation}.

\subsection*{B. Generalizable, robust performance}

When probing a particular ability, it is important to evaluate the models' performance across a variety of tasks, prompts, and scenarios. A failure to generalize beyond a particular surface-level form of the input may indicate that a model is using a non-human-like computational mechanism. For instance, it is sometimes the case that a model might perform well on a particular benchmark by leveraging low-level co-occurrence patterns (e.g., learning to predict that Sentence A entails Sentence B just because Sentences A and B have overlapping lexical items), but as soon as these obvious patterns are removed, the model's performance might drop to chance levels \cite{mccoy-etal-2019-right}. 
\vspace{.14in}

A particular danger when evaluating model abilities is excessive reliance on single examples. As in any scientific endeavor, assessing a phenomenon requires multiple observations to ensure generalizability and replicability. Thus, we here emphasize systematic benchmark-based evaluations rather than single examples (although those can be informative for illustrating a phenomenon or for initiating a more in-depth exploration). 
\end{boxed}
\end{figure*}

\subsection{Formal Reasoning}

Language allows people to discuss highly abstract ideas, turn ideas into scientific and philosophical theories, construct logical syllogisms, and engage in formal debates. Unsurprisingly, language is often considered a cornerstone of complex reasoning \cite{dennett_role_1994,carruthers_cognitive_2002}.
However, neuroscience provides evidence that language and formal reasoning dissociate in cognitive systems, and so a model that has mastered formal linguistic competence will not necessarily exhibit logical reasoning abilities.

\textbf{Humans.} Despite their close interplay, language and reasoning rely on distinct cognitive and neural systems. Unlike language, formal reasoning engages brain regions known as the \emph{multiple demand network} \cite{duncan_multiple-demand_2010}, named so because these regions are engaged in many cognitively demanding tasks: logic \cite{monti_thought_2012}, mathematical reasoning \cite{amalric_origins_2016}, physical reasoning \cite{fischer_functional_2016}, and computer code comprehension \cite{ivanovacode, liu2020computer}. Human patient studies have provided causal evidence for the role of the multiple demand network in logical reasoning by showing that the amount of damage to these regions correlates negatively with performance on standard tests of fluid intelligence \cite{woolgar_fluid_2010, woolgar2018}. Importantly, the multiple demand network supports reasoning even when the task is presented linguistically \cite{amalric_origins_2016, ivanovacode, monti_thought_2012} --- similar to how LLMs receive their prompts.

\textbf{LLMs.} Multiple studies have pointed out LLMs' limitations on tasks requiring formal reasoning, such as math problems. 
GPT-3 performs well on two-digit addition and subtraction but not on more complex tasks, such as three-digit addition or two-digit multiplication \cite{brown2020language}. GPT-4 similarly shows good performance on small-digit but not large-digit mathematical operations \cite{dziri}. Reasoning tests that break common co-occurrence patterns in the input or require multi-step operations also lead to model failure \cite{valmeekam2022large, wu2023reasoning}. 

The most commonly cited reason for these failures is the failure of artificial neural nets to generalize to patterns outside their training distribution \cite{wu2023reasoning,zhang}. This generalization gap can be partially bridged by ``chain of thought'' approaches, whereby a model is prompted to generate intermediate computation steps before arriving at an answer \cite{wei}. However, even these approaches do not lead to foolproof results \cite{dziri}. Thus, more and more researchers pair LLMs with external modules that can carry out structural logical and mathematical computations, such as the Mathematica plugin \cite{wolfram} or a probabilistic reasoning engine \cite{wong2023word}. The shift toward augmenting LLMs with reasoning-specific modules is consistent with evidence from neuroscience: language and formal reasoning are distinct cognitive capacities that work best when they are supported by separate processing mechanisms.

\subsection{World models 1: factual and commonsense knowledge} \label{world_knowledge}

A commonly debated capacity in LLMs is their ability to leverage internal world models \cite{yildirim2023task, wong2023word}. We break down the notion of world models into two components: world knowledge (factual and commonsense, this section) and situation tracking (the ability to maintain and update information about objects, agents, etc.; next section).

\textbf{Humans.} Evidence from neuroscience shows a dissociation between linguistic and semantic (world) knowledge. Individuals with language deficits may struggle to produce grammatical utterances and retrieve contextually appropriate words, but their ability to reason about objects and events presented non-linguistically often remains intact \cite{ivanova_language_2021, benn2023language}. On the other hand, individuals who suffer from semantic dementia (a neurodegenerative disorder) retain the ability to speak but struggle with tasks that rely on world knowledge (e.g., knowing that pumpkins are typically orange) even when the stimuli are presented non-verbally as pictures \cite{patterson_where_2007}. Thus, linguistic and semantic knowledge can be disentangled.

\textbf{LLMs.} LLMs have access to a wealth of knowledge about the world: word co-occurrence patterns in texts on the web contain both factual information (e.g., who was the first man on the moon) and commonsense information (e.g., the taste of lemon) \cite{grand2022semantic}. 
If this information can be effectively extracted, LLMs would be able to serve as off-the-shelf knowledge bases \cite{petroni2019language}. However, world knowledge contained in LLM representations suffers from several major shortcomings. 

First, LLMs routinely generate false statements, informally known as ``hallucinations''. This observation is unsurprising: their training objective is to generate plausible sentence continuations, with no reference to the underlying factual correctness of the resulting claims. Some developers have fine-tuned LLMs to provide links to sources that back up their claims; however, those citations can also be inaccurate \cite{liu-etal-2023-evaluating}.

Second, LLM outputs are often inconsistent: the same prompt phrased in different ways can elicit different responses \cite{sclar2023quantifying}. They can also get ``distracted'' by intervening information, e.g., an irrelevant claim inserted between a premise and a conclusion \cite{misra-etal-2023-comps}.

Third, commonsense knowledge is often underrepresented in language corpora: people are much more likely to communicate new or unusual information rather than commonly known facts \cite{gordon2013reporting}. As a result, LLMs can struggle on commonsense knowledge benchmarks \cite{liu-etal-2022-things}, especially once low-level statistical cues are controlled for \cite{elazar-etal-2021-measuring}.

And fourth, explicitly stated factual knowledge is easy to access but hard to maintain, requiring constant updates; for instance, the answer to ``Who is the current president of the US?'' will change every 4 or 8 years. Whereas humans can update their knowledge representations via a single sentence, updating world knowledge in LLMs requires locating and editing this particular bit of knowledge in their internal parameters-a non-trivial task \cite{kim2023carpe}, especially because these edits should affect some other bits of knowledge (e.g., that the previously current president is now the past president) but leave many other facts unaffected \cite{meng2022locating}. 

A more human-like approach to world knowledge representation might require dissociating linguistic representation/processing and world knowledge storage/updates. Such approaches exist \cite[e.g.,][]{borgeaud2022improving} but have not yet reached dominance in the field, typically because of relatively low coverage of existing knowledge bases. Although we cannot rely on LLMs alone for accurate world knowledge claims, we might use them as a starting point for constructing detailed knowledge bases \cite{cohen-etal-2023-crawling} and commonsense schemata \cite{chersoni2019structured}.

\subsection{World models 2: situation tracking} \label{situation-model}

People can follow the plot of a story that spans multiple chapters or even multiple books. We can also remember many details weeks or months after a conversation. We accomplish these feats by leveraging language inputs to create a "situation model" — a mental model of entities, relations between them, and a sequence of states they had been in or events they had participated in \cite{van}. 
Does the language network in humans construct a situation model based on its inputs? And how good are LLMs at building and updating situation models over time?

\textbf{Humans.} The language network in humans does not appear to track structure above the clause level \cite{lerner_topographic_2011, jacoby_discourse-level_2020}.
Instead, integration of meaning over longer periods of time likely takes place within the so-called default network \cite{buckner_brains_2019}.
Crucially, the default network tracks both linguistic and non-linguistic narratives \cite{baldassano_discovering_2017}, indicating that situation modeling is not a language-specific skill.

\textbf{LLMs.} Situation modeling in LLMs faces two main challenges: (1) extracting information from many sentences in a row; (2) integrating incoming inputs to appropriately update information about entities and their states.

The first problem is currently being tackled by continuously increasing the models' context window, i.e., the number of words they can process in one go. This approach will inevitably run into computational challenges: when summarizing a book, having a model that simultaneously attends to each word in that book is vastly inefficient (although see some attempts to overcome this issue, e.g., \cite{su2024roformer}). A human-like solution to this problem might include hierarchical processing, e.g., generating a summary for each chapter and then for the whole book (for related approaches, see \citealp{moirangthem2020abstractive, ruan-etal-2022-histruct}). 

Even when LLMs operate over shorter spans of text that easily fit inside their context windows, the question is: can they update their internal representations to track changes in the world? Some evidence suggests that they can \cite{kim-schuster-2023-entity}, although LLMs make characteristically non-human-like mistakes when it comes to situation modeling: for instance, their outputs can refer to non-existent discourse entities (``Arthur doesn't own a dog. The dog is brown.'' \cite{schuster-linzen-2022-sentence}). Thus, whether robust situation model building over shorter span of text is feasible using an LLM-only architecture remains a matter of debate.

\subsection{Social reasoning} \label{pragmatics}

``Water!''

Wittgenstein famously used single-word utterances like this to show that linguistic meaning radically depends on context. Although this word's literal meaning is straightforward, the intended meanings are more varied. Is the word being gasped by a thirsty person in the desert? By a hiker warning his friend of a hidden stream? An impatient diner talking to a waiter? Work in cognitive science and linguistics has come to recognize that these context-dependent aspects of language are not just peripheral but a central part of human language production and understanding \cite{clark1996using,grice_logic_1975}. 

The set of skills required to infer the intended meaning of an utterance beyond its literal content is known as pragmatics.
Pragmatics likely engages a variety of neural mechanisms \cite{andres-roqueta_contribution_2017, Levinson2000, hauptman2023non}, including both the language network and other brain regions. Thus, different types of pragmatic reasoning can be classified either as formal or as functional competence.
Here, we focus on one core functional competence capacity required for pragmatics: social reasoning.

\textbf{Humans.} A wealth of neuroscientific evidence shows that the human brain has dedicated machinery for processing social information \cite{deen_functional_2015, saxe_uniquely_2006}. The most relevant to our current discussion is the \textbf{theory of mind} network \cite{gopnik1992child}, a set of brain regions that are engaged when their owner is attempting to infer somebody’s mental state (with our without the use of language; \cite{saxe_people_2003, jacoby_localizing_2016}).
The specific contributions of the theory of mind network to language understanding can be divided into two broad categories. First, just like other functionally specialized brain modules, it is engaged when processing semantic content that is specifically related to its domain: narratives that require inferring the mental state of the characters engage the theory of mind network \cite{jacoby_localizing_2016}, and texts that require inferring the characters’ intentions evoke greater activity than those that do not \cite{ferstl_what_2002, saxe_its_2006}. Second, the theory of mind network is engaged more strongly during nonliteral language comprehension, including phenomena like jokes, sarcasm, indirect speech, and conversational implicature  \cite{hagoort_neuropragmatics_2014, hauptman2023non}---in other words, in situations where understanding the meaning of an utterance requires inferring the intentions of the speaker. Thus, successful language understanding relies on our broader, non-language-specific social inference skills.

\textbf{LLMs.} 
Recent models, trained with RLHF, have shown strong performance in interpreting non-literal utterances, such as metaphors and 
polite deceits, suggesting they can reach human or near-human performance on at least some pragmatic tasks \cite{hu-etal-2023-fine}. 
That said, LLMs exhibit uneven performance across pragmatic domains: their ability to interpret sarcasm or to complete jokes was limited even as their metaphor comprehension abilities soared \cite{hu-etal-2023-fine}. Overall, at least some forms of pragmatic inference might be acquired via targeted fine-tuning. It remains an open question whether aspects of pragmatics that are easiest for LLMs are those that are supported by the language network in humans.

LLMs' ability to solve theory of mind tasks has been subject to particular controversy. These tasks require both social knowledge and the ability to maintain a situation model. A typical example would feature character X moving an object from location A to location B while character Y is not around and so, does not see the move. The goal is to predict the true location of the object (location B) and the location where character Y believes the object is (location A). A bold claim that instruction-tuned LLMs have mastered theory-of-mind tasks \cite{kosinski2023theory} was quickly countered by a demonstration that including basic controls (such as character Y being told about the true object location) brought LLM performance to below-chance levels \cite{ullman2023large}. Several other studies have identified limitations in LLM performance on theory of mind tasks \cite[cf. \citealp{gandhi}]{shapira2023clever, sap-etal-2022-neural, trott2023large}. One solution to overcome these limitations has been to augment an LLM with a symbolic tracker of entity states and character beliefs \cite{sclar-etal-2023-minding}, an approach that mirrors the separation between language and theory of mind processing in humans.

\subsection{Language input can bootstrap functional competence capabilities}
Many non-linguistic cognitive capabilities can be substantially enhanced by language input. In humans, this relationship is particularly salient during development: babies learn new conceptual categories more easily when they are accompanied by linguistic labels \cite{waxman2002early}, and children with delayed language access have delayed social reasoning abilities \cite{pyers2009language}. Even in adulthood, knowledge of specific number words predicts the ability to conceptually represent exact numbers \cite{pitt2022exact}. Coupled with the fact that language inputs contain vast quantities of information about the world, and that language is both a crucial data source and representational substrate for much of people's world knowledge, this evidence suggests that, in principle, a model trained exclusively on language input could acquire much of functional linguistic competence.

Thus, we do not argue that functional linguistic competence is out of reach for language-based models; our main goals are (1) to highlight the conceptual distinction between formal and functional linguistic competence—which in the human brain draw on separate neural circuits, and (2) to demonstrate the gulf between LLMs’ formal and functional linguistic abilities. These facts lead to a speculation that, like the human brain, models that can master language use would also require or benefit from separate mechanisms for formal and functional competence. We discuss this idea next.

\section{Toward models that use language like humans} \label{future}

In this paper, we have advanced the thesis that formal and functional linguistic competence are distinct capabilities, with formal competence relying on distinct language machinery and function competence requiring the integration of diverse brain networks. We have shown that formal competence emerges in contemporary LLMs as a result of the word-in-context prediction objective; however, this objective alone appears insufficient for equipping LLMs with functional linguistic competence skills. Based on the neuroscience evidence, we suggest that models that succeed at real-life language use will need to be \emph{modular}, mimicking the division of labor between formal and functional competence in the human brain.

We see at least two ways to separate LLM circuits responsible for formal and functional competence: explicitly building modularity into the architecture of the system (we call this \textbf{Architectural Modularity}) or naturally inducing modularity through the training process, both through the training data and the objective function (we call this \textbf{Emergent Modularity}).

{Architectural Modularity} has a long history; it involves stitching together separate components, perhaps with quite specialized architectures \cite{bottou_framework_1990, ronco_neural_1997}. Modern-day examples include a transformer language model paired with a separate memory module \cite[e.g., ][]{borgeaud2022improving, liu_relational_2022} or a model for visual question answering, which includes a language module, a vision module, and a reasoning module \cite{mao2018the, hudson2019learning}. Such modular models achieve high task performance, are more efficient (i.e., can be trained on smaller datasets and have lower compute demands during inference), and show better generalizability (i.e., perform well on datasets with previously unseen properties). The modules of such models can be trained separately or together, similarly to how humans can flexibly combine different cognitive skills when learning to perform novel complex tasks.

Recently, the desire for this kind of modularity has expanded to include attempts to augment language models with the ability to call separate programs, as in including API calls \cite{schick2023toolformer}, mathematical calculators \cite{cobbe2021training}, planners \cite{liu2023llm}, and other kinds of modules that do specific structured operations.

Another approach in this vein uses LLMs as modules to translate a natural language query into code, which can then be passed to a symbolic module, which then generates an answer. \cite{wong2023word} outline a research program for this approach, showing that a version of GPT-3 fine-tuned to generate both natural language and code (Codex) can translate text input into meaningful structured probabilistic programs; inference in these programs can be used to reason over relational domains (like kinship systems), grounded domains (like visual scenes), and situations that require planning and understanding the plans of others. 
Their approach demonstrates a promising avenue for integrating what LLMs succeed at (namely, formal linguistic competence) with other cognitive modules that benefit from symbolic structure and abstraction.

The {Emergent Modularity} approach involves training models end-to-end (similarly to contemporary LLMs) while creating the conditions that facilitate the emergence of specialized model sub-components over the course of training. Modular structure has been shown to spontaneously emerge in some end-to-end neural network systems in domains other than language \cite[e.g., ][]{yang_task_2019, dobs_brain-like_2022}, which suggests that emergent modularity may constitute an optimal solution to many complex tasks. One strategy for this approach to be successful is for the model architecture to incentivize the development of individual, specialized modules within the model. Transformers, the most popular architecture today, satisfy this condition to some extent by allowing different attention heads to attend to different input features \cite[e.g.][]{manning2020emergent,vaswani2017attention,vig-belinkov-2019-analyzing}; certain approaches promote modularization even more explicitly, e.g., by endowing transformers with a mixture-of-experts architecture that incentivizes separate ``experts'' to carry out different computations \cite{goyal2022coordination, kudugunta-etal-2021-beyond-distillation, zhou2022mixture}. 

A modular model architecture is much better aligned with the brain's functional architecture for language, which includes separate components for formal and functional competence. Is it possible to build formally and functionally competent systems that do not mimic the modular structure of the human brain? In theory, yes: systems with different underlying architectures (e.g., modular vs. non-modular) can, in principle, exhibit similar behaviors. However, explicitly disentangling formal and functional competence skills at the architectural level is perhaps the most fail-safe path toward ensuring that an AI model uses language in a humanlike way.

\begin{boxed}[The need for better benchmarks] \label{box:benchmarks}
To assess progress on the road toward building models that use language in human-like ways, it is important to develop benchmarks that evaluate both formal and functional linguistic competence. This distinction can reduce the confusion that arises when discussing these models by combating the ``good at language -> good at thought'' and the ``bad at thought -> bad at language'' fallacies.

\vspace{.14in}

Several existing benchmarks already evaluate formal linguistic competence in LLMs \cite{gauthier2020syntaxgym,warstadt2020blimp} and can be complemented by additional tests of core linguistic features: hierarchy and abstraction. Benchmarks for evaluating different domains of functional linguistic competence, like commonsense world knowledge (e.g., WinoGrande \cite{sakag}), can often be ``hacked'' by LLMs by leveraging flawed heuristics \cite{elazar-etal-2021-back}. This issue is likely exacerbated in large-scale heterogeneous datasets like BIG-bench \cite{srivastavaBIG}. Moreover, functional competence benchmarks often rely on a certain, often underspecified level of required formal linguistic competence skills and/or mix together different functional competence abilities. Designing benchmarks that carefully disentangle different components of language knowledge and use would therefore constitute an important step toward a more informative assessment of LLMs.

\end{boxed}

\section{Concluding remarks}\label{conclusion}

Over the last few years, the discourse around language models has consisted of a curious mix of overclaiming and underclaiming \cite{bowman-2022-dangers}. 
While some claim models are on the verge of intelligence, others have pointed out the many failures of LLMs on a broad range of tasks, from number multiplication to generating factually true statements.
Here, we have put these seemingly inconsistent reactions in dialog with prior and ongoing work in computational linguistics, cognitive science, and neuroscience. In particular, we argue that LLMs are remarkably successful on tasks that require a particular type of structural and statistical linguistic competence—formal linguistic competence. Although their performance is not yet fully human-like, these models achieve an impressive degree of success in representing and using hierarchical relationships among words and building representations that are sufficiently abstract to generalize to new words and constructions. As such, these LLMs are underused in linguistics as candidate models of human language processing.

We also review some of the LLMs’ failures on tasks that target real-life language use, such as reasoning, while highlighting that the capabilities these tasks require are fundamentally distinct from formal language competence and rely on machinery in the human brain distinct from language processing machinery.

The failures of LLMs on non-linguistic tasks do not undermine their utility as models of language processing. After all, the brain areas that support language processing in humans also cannot do math, solve logical problems, or even track the meaning of a story across sentences or paragraphs. If we take the human mind and brain—a good example of generalized intelligence—as a guide, we might expect that future advances in developing intelligent systems will require combining language models with models that represent abstract knowledge and support complex reasoning, rather than expecting a single model (trained with a single word prediction objective) to do it all. Finally, to detect and monitor such advances, we need benchmarks that cleanly separate formal and functional linguistic competence (Box \ref{box:benchmarks}).
Formally and rigorously evaluating functional competence in LLMs will be informative for both science and engineering (see Outstanding Questions).

To those who have argued that most interesting aspects of human language cannot be learned from data alone, we say that LLMs compellingly demonstrate the possibility of learning complex syntactic features from linguistic input (even if, as of now, much more input is required than a typical child gets exposed to). To those who criticize LLMs for their inability to do complex arithmetic or to reason about the world, we say, give language models a break: given a strict separation of language and non-linguistic capabilities in the human mind, we should evaluate these capabilities separately, recognizing successes in formal linguistic competence even when non-linguistic capabilities lag behind. Finally, to those who are looking to improve the state of machine learning systems, we suggest that, instead of, or in addition to, continuously scaling up the models \cite{kaplan2020scaling}, more promising solutions will come in the form of modular architectures—built-in or emergent—that, like the human brain, integrate language processing with additional systems that carry out perception, reasoning, and action.

\section{Glossary}

\begin{itemize}
	\item \textbf{Abstraction} is, for our purposes, a linguistic representation that allows for generalization. Part-of-speech is one such example: words like ``dog'' and ``cat'' belong to the abstract category of ``nouns''.
 \item \textbf{Architectural Modularity} involves explicitly building distinct modules into a computational model, with each module responsible for achieving different goals.
 \item \textbf{Fine-tuning} is a process by which, after a model is pretrained, it receives additional training on new data, often for a specific purpose.
	\item \textbf{Formal linguistic competence} is the ability to get the \textit{form} of language right. It includes knowledge of word formation (e.g, phonology and morphology), knowledge of word meaning, and knowledge of rules and statistical patterns for how words combine to create sentences. Note that our use of the term `competence' differs from the classic competence/performance distinction in linguistics, given that in both models and humans separating competence and performance is often difficult.
  \item \textbf{Emergent Modularity} refers to the natural induction of modularity through the model training process, without explicitly building it into the architecture.
	\item \textbf{Functional linguistic competence} is the ability to use language to accomplish things in the world. It relies on a host of non-language-specific cognitive domains like formal reasoning, world knowledge, situation tracking, and social cognition.
	\item \textbf{Hierarchical structure} is a crucial property of language that allows it to be more than just a linear sequence of words. Rather, how words go together in a sentence is better captured by a tree-like structure, where some words and phrases are nested inside larger phrases.
  	\item \textbf{The language network} is the interconnected set of brain regions that respond selectively to language but not to non-linguistic inputs and tasks.
	\item \textbf{Large Language Models (LLMs)} 
	are models based on deep neural architectures (often but not always transformers) and trained on massive amounts of text using a word-in-context prediction task (sometimes with additional training objectives incorporated during or after the main training process). The term ``large'' refers to the number of parameters in these models, which ranges from millions to billions, as well as the size of the training data.
 \item \textbf{Pretraining} is the process by which a model is first trained on a general task (for LLMs, typically a text prediction task) before being trained or used for a more specialized purpose.
\item \textbf{Reinforcement Learning from Human Feedback (RLHF)} is a process by which reinforcement learning techniques are used to impart human preferences (e.g., as to which of two model outputs is preferred) to a model. It seems to lead to significant improvements on functional tasks.
\item \textbf{Theory of mind} is a cognitive skill that enables thinking and reasoning about the minds of others (i.e., what others know, believe, want, etc.).
\item \textbf{Tokens} are the basic units in language models. In earlier language models, they were often words or morphemes. In today's LLMs, they are often inferred from large amounts of text using an algorithm like Byte Pair Encoding. They can resemble words and morphemes, but sometimes also for subword or linguistically unnatural units.
\end{itemize}

\input{highlights.tex}

\input{outstanding_questions}

\section*{Acknowledgements}

For helpful conversations, we thank Jacob Andreas, Alex Warstadt, Dan Roberts, Kanishka Misra, students in the 2023 UT Austin Linguistics 393 seminar, the attendees of the Harvard LangCog journal club, the attendees of the UT Austin Department of Linguistics SynSem seminar, Gary Lupyan, John Krakauer, members of the Intel Deep Learning group, Yejin Choi and her group members, Allyson Ettinger, Nathan Schneider and his group members, the UT NLL Group, attendees of the KUIS AI Talk Series at Koç University in Istanbul, Tom McCoy, attendees of the NYU Philosophy of Deep Learning conference and his group members, Sydney Levine, organizers and attendees of the ILFC seminar, and others who have engaged with our ideas. We also thank Aalok Sathe for help with document formatting and references.

\section*{Funding Sources}

KM acknowledges funding from NSF Grant 2104995. AI was supported by funds from the Quest Initiative for Intelligence. EF was supported by NIH awards R01-DC016607, R01-DC016950, and U01-NS121471 and by research funds from the Brain and Cognitive Sciences Department, McGovern Institute for Brain Research, and the Simons Foundation through the Simons Center for the Social Brain.

\section*{Conflicts of Interest}

The authors declare no Conflicts of Interest.

\bibliography{references,anthology,refs_2022_Anya,aann}  

\end{document}

%% file: highlights.tex
\section{Highlights}

\begin{itemize}
\item Formal linguistic competence (getting the form of language right) and functional linguistic competence (using language to accomplish goals in the world) are distinct cognitive skills.
\item The human brain contains a network of areas that selectively support language processing (formal linguistic competence), but not other domains like logical or social reasoning (functional linguistic competence).
\item In the late 2010’s, Large Language Models trained on word prediction tasks began achieving unprecedented success in formal linguistic competence, showing impressive performance on linguistic tasks that likely require hierarchy and abstraction.
\item Consistent performance on tasks requiring functional linguistic competence is harder to achieve for Large Language Models and often involves augmentations beyond next word prediction.
\item Evidence from cognitive science and neuroscience can illuminate the capabilities and limitations of Large Language Models and pave the way toward better, human-like models of both language and thought.
\end{itemize}

%% file: outstanding_questions.tex
\section{Outstanding Questions}

\begin{itemize}
\item \textbf{How much functional competence can be acquired from the linguistic signal?} Humans use language as a substrate for knowledge, and so LLMs acquire non-linguistic information from the linguistic signal. How much information in this signal can be used to bootstrap functional competence? Are there aspects of functional competence that cannot be learned from language at all?
\item \textbf{How can we train competent language models on smaller amounts of data?} LLMs have achieved remarkable linguistic competence but they are trained on data very unlike what human children encounter. Although LLMs receive vastly more words (several orders of magnitude), they lack the richly structured and interactive input thought to be essential to child language acquisition. Would benefits emerge from training models in more interactive and human-like ways?
\item \textbf{Will the formal competence successes of LLMs transfer to other world languages?} Most LLM evaluations have taken place in English and a handful of other world languages. Building models for lower-resourced languages and evaluating them on both formal and functional dimensions is an important ongoing project.
\item \textbf{How long will the LLM growth continue}? 10 years ago, most researchers in the field would not have predicted that LLMs would be as advanced as they are today. Will current AI approaches lead to further revolutionary achievements in language and thought or would the mastery of functional linguistic competence require radically new approaches?
\item \textbf{Which is more promising: architectural modularity or emergent modularity?} If we want to build human-like modular systems, will it require explicitly building in functionally distinct components or can they be induced through end-to-end fine-tuning, e.g. with Reinforcement Learning from Human Feedback (RLHF)?
\item \textbf{How modular are today's LLMs?} We argue that an LLM that achieves both formal and functional competence may need to rely on separate mechanisms for different competence types. Mechanistic interpretability studies can shed light on the extent to which different cognitive tasks might segregate even within today's LLMs.
\item  \textbf{Should LLMs be described as individual language users or as distributions over potential user outputs?} There are different ways to think about LLMs in the context of language use: e.g., as individual language users (``agent-based view'') or as tools augmenting human activities, like a calculator (``tool-based view'') \cite{yiu2023transmission,lederman2024language,mitchell2023debate}. Which of these views will be the most fruitful way to think about LLMs as they gain additional competencies and become more widely used?
\item \textbf{How much grounding do LLMs need to continue improving?} How limited are text-only approaches, compared to approaches across different modalities  \citep{pavlick2023symbols,cm}?
\item \textbf{How much can LLMs ultimately tell us about human language and cognition?} What are the cases where LLMs fall short as models of human language use? Are these discrepancies solvable or will they require language researchers to develop different paradigms?
\end{itemize}